%% file: main.tex
\documentclass[sigconf]{acmart}
\AtBeginDocument{%
  }

\setcopyright{acmlicensed}
\copyrightyear{2018}
\acmYear{2018}
\acmDOI{XXXXXXX.XXXXXXX}
\acmConference[Conference acronym 'XX]{Make sure to enter the correct
  conference title from your rights confirmation email}{June 03--05,
  2018}{Woodstock, NY}
\acmISBN{978-1-4503-XXXX-X/2018/06}

\usepackage{enumitem}
\usepackage{xcolor} 
\usepackage{tabularx}
\usepackage{subcaption}
\usepackage{booktabs}
\usepackage{multirow}
\usepackage{makecell}
\usepackage{array}

\begin{document}

\title{How Uncertain Is the Grade? A Benchmark of Uncertainty Metrics for LLM-Based Automatic Assessment}


\author{Hang Li}
\authornote{Both authors contributed equally to this research.}
\email{lihang4@msu.edu}
\affiliation{%
  \institution{Michigan State University}
  \city{East Lansing}
  \country{USA}
}

\author{Kaiqi Yang}
\authornotemark[1]
\email{kqyang@msu.edu}
\affiliation{%
  \institution{Michigan State University}
  \city{East Lansing}
  \country{USA}
}

\author{Xianxuan Long}
\email{longxia2@msu.edu}
\affiliation{%
  \institution{Michigan State University}
  \city{East Lansing}
  \country{USA}
}

\author{Fedor Filippov}
\email{filipo2@msu.edu}
\affiliation{%
  \institution{Michigan State University}
  \city{East Lansing}
  \country{USA}
}

\author{Yucheng Chu}
\email{chuyuch2@msu.edu}
\affiliation{%
  \institution{Michigan State University}
  \city{East Lansing}
  \country{USA}
}

\author{Yasemin Copur-Gencturk}
\email{copurgen@usc.edu}
\affiliation{%
  \institution{University of Southern California}
  \city{Los Angeles}
  \country{USA}
}

\author{Peng He}
\email{peng.he@wsu.edu}
\affiliation{%
  \institution{Washington State University}
  \city{Pullman}
  \country{USA}
}

\author{Cory Miller}
\email{mill3118@msu.edu}
\affiliation{%
  \institution{Michigan State University}
  \city{East Lansing}
  \country{USA}
}

\author{Namsoo Shin}
\email{namsoo@msu.edu}
\affiliation{%
  \institution{Michigan State University}
  \city{East Lansing}
  \country{USA}
}

\author{Joseph Krajcik}
\email{krajcik@msu.edu}
\affiliation{%
  \institution{Michigan State University}
  \city{East Lansing}
  \country{USA}
}

\author{Hui Liu}
\email{liuhui7@msu.edu}
\affiliation{%
  \institution{Michigan State University}
  \city{East Lansing}
  \country{USA}
}

\author{Jiliang Tang}
\email{tangjili@msu.edu}
\affiliation{%
  \institution{Michigan State University}
  \city{East Lansing}
  \country{USA}
}

\renewcommand{\shortauthors}{Li et al.}

\begin{abstract}

\input{abstract}
\end{abstract}


\input{cssxml}


\received{20 February 2007}
\received[revised]{12 March 2009}
\received[accepted]{5 June 2009}

\maketitle

\section{Introduction}
\label{sec:intro}
\input{introduction}

\section{Related Work}
\input{related}

\section{Method}
\input{method}

\section{Evaluation}
\input{evaluation}

\section{Conclusion}
\input{conclusion}


\bibliographystyle{ACM-Reference-Format}
\bibliography{custom-ref}










\end{document}

%% file: abstract.tex
The rapid rise of large language models (LLMs) is reshaping the landscape of automatic assessment in education. Benefiting from strong prior knowledge and advanced reasoning capabilities, LLM-based assessment systems have moved beyond laboratory prototypes used by small groups of researchers and are increasingly becoming practical tools for everyday use by teachers. While these systems demonstrate substantial advantages in adaptability to diverse question types and flexibility in output formats, they also introduce new challenges related to output uncertainty, stemming from the inherently probabilistic nature of LLMs. Output uncertainty is an inescapable challenge in automatic assessment, as assessment results often play a critical role in informing subsequent pedagogical actions, such as providing feedback to students or guiding instructional decisions. Unreliable or poorly calibrated uncertainty estimates can lead to unstable downstream interventions, potentially disrupting students’ learning processes and resulting in unintended negative consequences. To systematically understand this challenge and inform future research, we benchmark a broad range of uncertainty quantification methods in the context of LLM-based automatic assessment. Although the effectiveness of these methods has been demonstrated in many tasks across other domains, their applicability and reliability in educational settings, particularly for automatic grading, remain underexplored. Through comprehensive analyses of uncertainty behaviors across multiple assessment datasets, LLM families, and generation control settings, we characterize the uncertainty patterns exhibited by LLMs in grading scenarios. Based on these findings, we evaluate the strengths and limitations of different uncertainty metrics and analyze the influence of key factors, including model families, assessment tasks, and decoding strategies, on uncertainty estimates. Our study provides actionable insights into the characteristics of uncertainty in LLM-based automatic assessment and lays the groundwork for developing more reliable and effective uncertainty-aware grading systems in the future.

%% file: cssxml.tex
\ccsdesc[500]{Computing methodologies~Natural language processing}
\ccsdesc[500]{Applied computing~Computer-assisted instruction}

%% file: introduction.tex
With the surge of Large language models (LLMs), LLM-powered automatic assessment has become increasingly popular in many domains~\cite{gu2024survey,Li2024FromGT}. Among the scenarios employing LLMs as assistants, automatic assessment is one of the most practical and challenging tasks. Automated assessment, also called automated grading, is a system that takes questions and candidate solutions, optionally with rubrics and other context information, and generates text to present the correctness of the evaluated solution~\cite{ala2005survey,Emirtekin2025LargeLM}. The inputs of automated assessment include questions, answers, rubrics that describe how to assess the solutions, and optionally, demonstrations of solution-grade examples. From the output side, assessment results are not limited to values or labels of grades, but also cover aspect-level assessment~\cite{Low2019AutomatedAO,Song2024FineSurEFS}, reasons for assessments~\cite{Cohn2024ACP,Li2024AutomatedFF}, etc. In existing work, the automated assessments are supported by rule-based~\cite{braun2006rule,cline2010rule} or supervised machine learning models~\cite{ahmed2022deep,lu2021integrating}, which require intensive labor from domain experts to design the rules and features, to annotate large-scale datasets for training and validation, and to address anomalies or out-of-distribution cases~\cite{ala2005survey, paiva2022automated}. Owing to LLMs' broad prior knowledge and instruction-following ability, LLM-based grading is able to deal with diverse question domains, understand rubrics, and produce structured evaluations together with interpretable rationales in natural language~\cite{Li2025LLMBasedAG, Kortemeyer2023PerformanceOT, Hashemi2024LLMRubricAM}. Compared with traditional task-specific grading systems, LLM-based systems offer a unified pipeline capable of handling diverse inputs without training or adaptation. This flexibility has enabled their usage across domains and problem formulations in practical educational workflows.

Although automated assessment is efficient, the uncertainty issue has been a central challenge, and this is even more severe in LLM-based assessment. Uncertainty is the metric denoting how sure the models are when making judgments, with the causes of randomness, absence of precise boundaries of concepts, inaccurate base knowledge, and ambiguity between different objects~\cite{mena2021survey}. Uncertainty causes problems of decreased performance and misleading outputs, especially for cases without adequate information or model abilities. This issue is more severe in LLM-based frameworks, as LLMs hide the internal mechanisms in large-scale parameters, making it hard to probe or control. In addition, LLMs rely on token probability and sampling, which inherently brings randomness into the generation. Lastly, being powerful in text generation, LLMs tend to produce seemingly correct responses even without enough information and capabilities, which further increases the risk of misleading results. In conclusion, certainty is highly prioritized by automated assessment tasks, and responses of low certainty are harmful to both the learning procedures and the equality between students. 

Uncertainty quantification (UQ) has been studied in various tasks, including classification~\cite{kurz2022uncertainty, chlaily2023measures}, regression~\cite{levi2022evaluating, abadie2020sampling}, and Bayesian inference settings~\cite{mena2021survey}, covering data in text and image modalities. However, the existing methods rely heavily on access to model structure and hidden states, including prediction probabilities represented by neural networks~\cite{sensoy2018evidential}, the feasible space of possibility distributions~\cite{chlaily2023measures}, etc. Although showing convincing performances in measuring uncertainty, the existing methods are inapplicable for LLM-based auto-assessment systems, because the internal states of LLMs are either computationally expensive and hard to handle, or even inaccessible when using proprietary LLMs for users. There are also works on uncertainty estimation for LLMs, which primarily include white-box methods that require LLMs to provide logits, internal layer outputs, and the model parameters~\cite{vashurin2025benchmarking}; consistency-based methods that ground uncertainty on the comparison between multiple generations~\cite{zhang2024luq}; and reflection-based methods that elicit LLMs to self-report or analyze the certainty in generation~\cite{tian2023just}. Because the access to models and settings of the methods is different, these works are not systematically compared, and there are few insights into how well they perform. As a prosperous research field, LLM-based auto-assessment has developed many works on modeling and implementing uncertainty, while a comprehensive investigation to validate and compare them is still ignored. 

To bridge this gap, in this work, we propose a benchmark on uncertainty quantification for the auto-assessment tasks. The challenges of applying UQ methods to auto-assessment derive from the specificity in the wide range of domain knowledge, the finite set of ordinal output features, and the influences of input prompts. First, auto-assessment deploys general-use LLMs as a judge in a specialized domain, which requires benchmarking work to consider heterogeneous subjects, rather than being limited to domains of LLMs' expertise. Second, the expected output (i.e., the grade) is an ordinal variable whose labels have a relative order, but the distances between categories are not comparable. This prevents arithmetic operations on the grade labels and makes numeric metrics (e.g., differences and mean squared error) invalid. Similarly, metrics designed for semantic distance may not adequately capture the differences of outputs, because both the correct and wrong gradings are centered on the semantic representations of question contexts. Third, LLMs are sensitive to the input text, making many UQ methods incomparable as they use different templates along with the question and rubrics. 

Taking these points in mind, we design the settings of the benchmark with multiple learning subjects, different forms of output grades, and unified templates to build the input prompts from questions and rubrics. We collect datasets of both open-sourced and private, covering essay writing, natural science, chemistry, and math education. As for the output grades, the grade sets include binary, ordered, multi-class, etc. In addition, we elicit the LLMs to judge with the text of reasons for the grades, incorporating the semantic features into the analysis. Finally, we use a set of curated-designed templates to build the prompts, trying to make the input text similar to prior works or the other datasets in this benchmark. To satisfy the need for broad usage and flexibility, we focus on the repetition-based (or ensemble-based) methods in this work, which take several runs of generation to estimate uncertainty, without using internal states of LLMs in any way. 

%% file: related.tex
\subsection{LLM-based Automatic Assessment}
Automatic assessment has long been studied before the emergence of LLMs. Earlier systems were typically built on task-specific machine learning models trained for particular grading formats, such as automated essay scoring, short-answer evaluation, or rubric-based classification~\cite{ala2005survey,paiva2022automated,higgins2005automated}. These approaches generally relied on engineered linguistic features and supervised neural architectures for fixed domains or tasks. While effective in narrow settings, they often require annotated datasets for each task, making it costly to scale assessment systems to new contexts. Besides, they require rigorous prompt tuning to align the assessment with fine-grained evaluation criteria, and this limits the transferability across subjects, question types, and scoring criteria.

The advent of large language models has shifted this paradigm toward general-purpose evaluators that can assess with instructions and rubrics. A representative framework is {LLM-as-a-Judge}~\cite{gu2024survey}, where a prompted model assigns correctness labels~\cite{zhu2023judgelm,liu2023g,gao2023human} or reviews written works~\cite{brake2024comparing,hosseini2024benchmark}. The LLM-as-a-Judge framework is widely used to handle diverse tasks, including mathematics~\cite{luo2023wizardmath}, natural and medical science~\cite{zhao2024artificial,krolik2024towards}, programming~\cite{kumar2024llms}, and language~\cite{song2024automated}. Recent studies demonstrate the growing capability of LLM-based assessment systems. For example, LLM evaluators can produce stable grading judgments across diverse prompts and reasoning settings~\cite{raina2024llm}; rubric-guided prompting enables models to perform fine-grained scoring aligned with evaluation criteria~\cite{deng2025rubric}; aggregation frameworks such as SURE~\cite{korthals2025towards} leverage self-consistency and majority voting to improve grading reliability. Along with prompt engineering work, there are also studies on how to refine rubrics to achieve fine-grained grades~\cite{singh5525975optimizing,chu2024llm}. However, these studies also reveal remaining limitations: evaluation outcomes may change under minor input perturbations, including the replacement of synonyms and meaning-invariant rephrasing~\cite{Zhu2023PromptRobustTE,raina2024llm,He2024DoesPF}, symbols indicating grade labels~\cite{Ye2024JusticeOP}, and the order of provided grade labels~\cite{Li2025EvaluatingSB,Shi2024JudgingTJ}. These findings suggest that, although LLM-based assessment improves coverage and flexibility, existing methods remain vulnerable to undesired variation in generated judgments. This limitation highlights the need for uncertainty estimation to complement assessment decisions and mitigate unreliable outputs.

\subsection{Uncertainty Estimation in LLMs}
With the surge of LLM applications, inaccurate or non-factual generations have motivated uncertainty estimation as a practical signal of when outputs should (or should not) be trusted. A common starting point is to derive confidence from {observable generation behavior}, especially via repeated sampling. \cite{cole2023selectively} shows that repetition-based metrics computed from multiple samples provide a reliable criterion for abstention decisions against uncertain responses, outperforming likelihood-based and self-verification alternatives. \cite{lyu2024consistency} formalizes sample consistency into confidence measures based on agreement or entropy, and demonstrates their effectiveness for confidence calibration across models and reasoning tasks. This work also explores how factors such as sampling size and explanation prompting influence calibration quality. In addition, \cite{xiong2023can} proposes self-probing, a method that elicits verbalized confidence from the LLMs, and finds that the verbalized confidence score helps with calibration and failure prediction, while also presenting overconfident statements like humans. Complementing the method-focused studies, ~\cite{huang2023look} conducts a large-scale empirical comparison with a broad set of uncertainty estimators, LLMs, and domain tasks, claiming that uncertainty signals can help surface risky generations while also exposing practical limitations across settings. Beyond focusing on the match of surface features, more sophisticated approaches refine the notion of {equivalence} used to aggregate samples. For open-ended generation, \cite{kuhn2023semantic} introduces semantic entropy, which clusters responses by meaning invariance and quantifies uncertainty over semantic modes rather than token sequences, thereby addressing the mismatch between lexical diversity and true epistemic uncertainty. Recently, \cite{xia2025survey} synthesizes inference-time uncertainty estimation for LLMs by defining key uncertainty sources (e.g., incomplete information and model limitations) and organizing methods into a taxonomy that includes verbalized confidence, latent-information, consistency-based, and semantic clustering approaches.

In addition, the ordinal characteristic of grade labels makes the assessment task different from other tasks. The grades are presented with a label or value that belongs to a finite set in most cases; for example, in SemEval 2013 - Task 7~\cite{dzikovska2013semeval} where the grade labels indicate whether and for which degree student answer aligns with the given reference, the scores are: \texttt{3} for fully correct, \texttt{2} for partially correct, \texttt{1} for contradictory with reference, and \texttt{0} for irrelevant. Although the labels are inherently ordered in a sequence where higher scores indicate a stronger tie with the reference, the distance between labels is incomparable or even nonsensical~\cite{tellamekala2023cold,xie2024trustworthy}. This body of work motivates our benchmark design. When evaluating these uncertainty signals in automated assessment, we focus on grading tasks whose predictions lie in a constrained label space.

%% file: method.tex
\subsection{Uncertainty Definition}
\label{sec:uncertainty_definition}
As described in Section~\ref{sec:intro}, we focus on uncertainty estimation approaches based on repeated generation, as such methods can be applied uniformly to both open-source and proprietary LLMs without requiring access to internal model states. Following prior work~\cite{kuhn2023semantic,vashurin2025benchmarking}, uncertainty is defined as the variability in a model’s predictions. In the context of LLM-based automatic assessment, uncertainty can be operationalized as the variability of grading outputs $\mathcal{O}$ produced for a given input triplet $x = (q, r, a)$, consisting of the question, grading rubric, and student answer. Formally, the uncertainty associated with $x$ is defined as

\begin{equation}
    U = f(\mathcal{O}),
    \quad
    \mathcal{O} =\{ o_{i} \}_{i=1}^{N},
    \quad
    o_i = g_{\theta}(x, c)
\end{equation}

\noindent where $f(\cdot)$ denotes an uncertainty quantification function, $g_{\theta}$ is the grading model parameterized by $\theta$, $o_i$ denotes the grading output from the i-th repeated generation, $N$ is the repetition times, and $c$ represents additional contextual information incorporated by recent inference-enhancement techniques, such as few-shot demonstrations~\cite{zhao2025language} and external knowledge retrieved via retrieval-augmented generation (RAG)~\cite{chu2025enhancing}.

\subsection{Uncertainty Quantification Methods}

Existing studies on uncertainty quantification have primarily focused on open-ended question answering (Q\&A) or reasoning tasks~\cite{xia2025survey}, where model responses are free-form. Accordingly, many existing methods characterize uncertainty through measures of semantic equivalence or variability across generated responses. In this work, we shift the focus to uncertainty quantification in the context of automatic assessment. Unlike open-ended generation, automatic assessment is typically formulated as a classification or scoring task, in which model outputs are constrained to a fixed label or score space. Consequently, we select uncertainty quantification methods~\cite{xia2025survey,lyu2024consistency,kuhn2023semantic} that are suitable for capturing uncertainty from a categorical perspective. Meanwhile, with the recent adoption of chain-of-thought (CoT) prompting~\cite{wei2022chain} and the emergence of reasoning-oriented models (e.g., GPT-o1), model outputs in automatic assessment increasingly include not only final grading scores but also intermediate grading rationales. To capture uncertainty manifested in these intermediate representations, we additionally incorporate representation-based uncertainty quantification methods into our study. In the following sections, we present the uncertainty quantification methods in each category in detail.

\subsubsection{Categorical Based Methods} Owing to the pre-defined and discrete output label space, methods in this category are relatively simple to compute and computationally efficient. In general, they derive uncertainty from the empirical frequency distribution of score labels observed across repeated queries to the same request.

\paragraph{\textbf{Numset}~\cite{lyu2024consistency}} It measures uncertainty as the number of unique answers observed across multiple samples:

\begin{equation}
    U_{\mathrm{Numset}} = |\mathcal{O}_u|
\end{equation}

\noindent where $\mathcal{O}_u$ denotes the set of unique answers obtained from repeated queries to the same input $x$. Intuitively, a larger $|\mathcal{O}_u|$ indicates greater variability in the model’s outputs and thus higher uncertainty.

\paragraph{\textbf{Max-Agree-Rate (MAR)}~\cite{xiong2023can}} It captures uncertainty by incorporating the full distribution of answer frequencies. Compared to Numset, which only counts the number of distinct answers, the MAR reflects the degree of dominance of the most frequent answer among all sampled responses. A larger MAR indicates that the model consistently produces the same answer and is therefore more confident. To express this as an uncertainty measure, we define the MAR as

\begin{equation}
    U_{\text{MAR}} = 1 - \max_{o \in \mathcal{O}_u} \frac{1}{N}\sum_{i=1}^{N} \mathbf{1}[o_i = o],
\end{equation}

\paragraph{\textbf{Categorical-Entropy (CE)}~\cite{lyu2024consistency}} It is a classical distribution chaos quantification method over the output class probabilities for classification problems~\cite{de2005tutorial}. As the automatic assessment is also commonly treated as the classification task, we adopt the CE as the uncertainty evaluation method for the automatic assessment. Compared to the MAR, CE not only captures the value distribution of the majority class but also accounts for all categories, ensuring that shifts in the distribution of minority responses are reflected in the computed value. Following the common definition in classification, we define entropy-based consistency $U_{\rm{CE}}$ as:

\begin{equation}
    U_{\rm{CE}} = \sum_{o\in\mathcal{O}_u} - p_o\log(p_o),\ \ \ \ p_o=\frac{1}{N}\sum^N_i\mathbf{1}[o_i=o]
\end{equation}

\noindent where $p_o$ is the normalized frequency of each unique answer $o\in\mathcal{O}_u$. Commonly, a higher $U_{\rm{entropy}}$ indicates the more chaos existing in the distributions of the answers, which reflects a high uncertainty.  
\paragraph{\textbf{First–Second Distance (FSD)}~\cite{lyu2024consistency}} It is a recently proposed frequency-based uncertainty measure that aims to address limitations of commonly used methods such as MAR and Entropy. FSD quantifies the gap between the proportions of samples supporting the most frequent (majority) answer $\bar{a}$ and the second most frequent answer $\bar{\bar{a}}$. It is defined as:

\begin{equation}
U_{\rm{FSD}} = 1 - \frac{1}{N} \left( \sum_{i=1}^{N} \mathbf{1}[o_i = \bar{o}] - \sum_{i=1}^{N} \mathbf{1}[o_i = \bar{\bar{o}}] \right)
\end{equation}

Compared with the two prior methods, FSD mitigates the shortcomings of Entropy, which can be overly influenced by low-frequency tail categories, and of MAR, which depends solely on the majority answer and discards information about competing alternatives. By focusing on the relative dominance between the top two categories in the frequency distribution, FSD is particularly informative in cases where the model is uncertain between the two most plausible answers. In such scenarios, an FSD-based measure avoids overconfidence induced by considering only the most-voted answer. Intuitively, a larger gap between the top two frequencies indicates higher confidence in the majority answer. To align this measure with our uncertainty formulation (where larger values correspond to higher uncertainty), we use the complement of this gap, as defined above, as the uncertainty score.

\subsubsection{Relation Based Methods} 
Although evaluating uncertainty directly in the categorical label space is computationally convenient, it has notable limitations. Due to the discrete nature of label categories in automatic assessment tasks, estimated uncertainty values can exhibit large fluctuations when the number of sampled responses $N$ is small. For example, entropy-based measures (e.g., CE) are particularly sensitive to sampling noise before the response distribution stabilizes~\cite{cui2019class}. In practical automatic assessment settings, repeatedly querying an LLM for the same input $x$ is constrained by efficiency and cost considerations, making such instability especially problematic. Fortunately, the recent emergence of reasoning-path generation in large language models provides rich intermediate signals that help mitigate the limitations of category-based uncertainty quantification. Specifically, by modeling semantic similarity~\cite{reimers2019sentence} or logical entailment~\cite{kuhn2023semantic} among the reasoning traces associated with grading outputs ($o_i$ and $o_j$), and constructing relation graphs in which these quantified relationships serve as edge weights, relation-based uncertainty quantification methods enable more fine-grained and continuous assessments of model uncertainty. Existing relation-based approaches typically follow a common two-step paradigm: (1) constructing a relation graph over the set of model outputs, and (2) computing uncertainty metrics based on properties of the resulting graph. In the following, we describe these two steps in detail and present representative implementations.

\paragraph{\textbf{Relation Graph}}

To evaluate the uncertainty of a model’s responses, we conceptualize the problem as analyzing the “tightness” of relationships among outputs $o$ within a generated grading response set $\mathcal{O}$. Intuitively, a more compact and coherent group of answers indicates lower uncertainty, whereas a more dispersed group suggests higher uncertainty. Inspired by graph-based clustering analysis, recent studies~\cite{huang2023look,xia2025survey} construct a relation graph $\mathcal{G} = (V, E)$ over the outputs $\mathcal{O}$, where each vertex $v \in V$ corresponds to a grading response in $o\in\mathcal{O}$, and each edge $e_{ij} \in E$ is weighted by the pairwise distance between responses $o_i$ and $o_j$. A key factor influencing the analysis lies in how this relation graph is constructed: different definitions of pairwise relationships yield different notions of answer similarity and capture uncertainty at varying semantic depths and computational costs. In general, existing relation-based uncertainty quantification methods primarily differ in their choice of pairwise distance or similarity function. Below, we summarize three mainstream and representative approaches that we adopt for subsequent evaluation.

\begin{itemize}[leftmargin=1em]
    \item \textbf{Jaccard Similarity} measures lexical overlap between two answers based on the Jaccard distance between their token sets. Specifically, given two responses $a_i$ and $a_j$, the Jaccard similarity ($s_{ij}^{jac}$) is defined as the ratio between the size of the intersection and the size of the union of their token sets. 
    \begin{equation}
        s_{ij}^{\text{jac}} = \frac{|T(o_i)\cap T(o_j)|}{|T(o_i)\cup T(o_j)|},
    \end{equation}
    where $T(a)$ denotes the set of tokens in answer $a$. This approach captures surface-level similarity and is computationally efficient, but it is limited in its ability to reflect deeper semantic equivalence between paraphrased or lexically diverse answers.
    
    \item \textbf{Embedding Cosine Similarity} is defined based on the cosine similarity $s_{ij}^{\text{emb}}$ between vector representations of answers obtained from a sentence embedding model:
    \begin{equation}
        s_{ij}^{\text{emb}} = \frac{\mathbf{h}_i^\top \mathbf{h}_j}{\|\mathbf{h}_i\|_2 \, \|\mathbf{h}_j\|_2},
        \quad
        \mathbf{h}_i = f_{\text{emb}}(o_i),
    \end{equation}
    where $\mathbf{h}_i \in \mathbb{R}^d$ denotes the semantic embedding of answer $a_i$ produced by a pre-trained encoder $f_{\text{emb}}$. By operating in a continuous semantic embedding space, embedding-based similarity captures higher-level semantic relatedness beyond exact lexical overlap. Compared with \textbf{Jaccard Similarity}, this approach provides a more robust notion of answer similarity, at the cost of additional computational overhead for embedding extraction.
    
    \item \textbf{Entailment Score} models pairwise relationships using natural language inference (NLI) by estimating the degree to which one answer entails another. Off-the-shelf NLI models output class probabilities over entailment, neutral, and contradiction~\cite{lewis2020bart}. However, the limited input context window makes them unsuitable for directly processing long, multi-step reasoning traces produced by LLMs. To address this, we adopt a BERTScore-inspired strategy~\cite{zhang2019bertscore}: we split each answer into sentences and compute entailment scores over all sentence pairs, followed by a mean–max aggregation. Following prior work~\cite{kuhn2023semantic}, we use the entailment probability to construct the relation graph. Since entailment is directional, we symmetrize the pairwise relationship by averaging both directions:

    \begin{equation}
    s_{i\rightarrow j}^{\text{nli}}
    = \frac{1}{M} \sum_{m=1}^{M} \max_{k} P(\text{entail} \mid o_{i,m}, o_{j,k}),
    \
    s_{ij}^{\text{nli}}
    = \frac{1}{2}\big(s_{i\rightarrow j}^{\text{nli}} + s_{j\rightarrow i}^{\text{nli}}\big)
    \end{equation}

    where $P(\cdot)$ denotes the entailment probability predicted by the NLI model, M and K are the numbers of sentences after splitting answers $o_i$ and $o_j$, respectively, and $o_{i,m}$ denotes the m-th sentence of answer $o_i$, $o_{j,k}$ denotes the k-th sentence of answer $o_j$. Entailment scores capture fine-grained semantic and logical consistency between responses and are well-suited for identifying subtle disagreements in reasoning or factual content. While this method provides the most semantically grounded notion of relational tightness, it is also the most computationally expensive due to the need to run NLI inference over all sentence pairs.
    
\end{itemize}

\paragraph{\textbf{Property-driven Uncertainty}}

Given the relation graph $\mathcal{G}$, different graph-theoretic properties can be leveraged to quantify its tightness. Below, we describe four representative property-driven uncertainty measures that have been widely adopted in prior work on uncertainty quantification for Q\&A and reasoning tasks~\cite{xia2025survey}.

\begin{itemize}[leftmargin=1em]
    \item \textbf{Normalized Average Degree (NAD)}, also known as the variational ratio (VR)~\cite{huang2023look}, measures the connectivity density of the relation graph. Intuitively, if each answer is closely connected to many others, the response set forms a dense and compact cluster, indicating low uncertainty. Formally, we define the NAD-based uncertainty as

    \begin{equation}
        U_{\rm{NAD}} = 1 - \frac{1}{N(N-1)}\sum_{i=1}^{N}\sum_{i\neq j}s_{ij}^{*}
    \end{equation}

    where $s^{*}_{ij}$ denotes the pairwise similarity between answers $a_i$ and $a_j$ introduced in the previous section. Larger values of $U_{\mathrm{NAD}}$ indicate weaker local connectivity and thus higher uncertainty.
    
    \item \textbf{Graph Eccentricity (GE)} quantifies the maximum shortest-path distance from each node to all other nodes in the graph, and the overall tightness is summarized by aggregating node eccentricities. We define the eccentricity-based uncertainty as

    \begin{equation}
        U_{\rm{ECC}} = \frac{1}{N}\sum_i \max_j [d_{ij}^{\rm{sp}}], 
    \end{equation}

    where $d^{\rm{sp}}$ denotes the all-pairs shortest-path distance induced by the complement of similarity, $1-s_{ij}^*$. Smaller values indicate that all answers are mutually close in the relation graph, forming a compact cluster, whereas larger values imply that some answers are distant from the rest, reflecting higher uncertainty. GE thus characterizes the global spread of the response set.

    \item \textbf{Graph Eigenvalues (Eigen)} leverage spectral properties of the graph Laplacian to provide a global measure of connectivity. In particular, the second smallest eigenvalue, $\lambda_2(L)$, known as algebraic connectivity, reflects how tightly connected the graph is. We define the eigenvalue-based uncertainty as

    \begin{equation}
        U_{\mathrm{Eigen}} = \frac{1}{\lambda_2(L)}, \quad L = D - A, \quad A_{ij} = s^*_{ij}
    \end{equation}
    
    where $L$ is the graph Laplacian constructed from the degree matrix $D$ and adjacency matrix $A$. A more tightly connected graph exhibits larger $\lambda_2(L)$ and thus lower uncertainty, whereas smaller $\lambda_2(L)$ corresponds to higher uncertainty.

    \item \textbf{Discrete Semantic Entropy (DSE)} extends entropy-based uncertainty measures to the relation-graph setting by computing entropy over semantically clustered responses~\cite{kuhn2023semantic}. By grouping answers into equivalence classes based on pairwise relations in $\mathcal{G}$ and computing the entropy of the induced discrete distribution, DSE captures both the diversity of semantic modes and their relative frequencies. Specifically, we define

    \begin{equation}
        U_{\mathrm{DSE}} = - \sum_{m=1}^{M} p_m \log p_m,\quad p_m = \frac{|\mathcal{C}_m|}{N}
    \end{equation}
        
    where $\{\mathcal{C}_m\}_{m=1}^{M}$ denotes the clusters obtained from the relation graph $\mathcal{G}$. As clustering over $\mathcal{G}$ incurs additional computational cost, we follow prior work~\cite{kuhn2023semantic} and compute this metric only for the NLI-based relation graph, using bidirectional entailment relations. Here, $|C_m|$ is the size of clusters. Larger $U_{\rm{DSE}}$ indicates a more fragmented semantic landscape and thus higher uncertainty.
    
\end{itemize}

%% file: evaluation.tex
\subsection{Metric}

\subsubsection{Effectiveness}

As introduced in Section~\ref{sec:uncertainty_definition}, uncertainty values are designed to indicate the confidence or reliability of responses produced by LLMs. In practice, we expect responses with lower uncertainty to exhibit higher accuracy than those with higher uncertainty. In the context of automatic assessment, this property enables uncertainty to serve as a signal for human intervention, preventing low-reliability grading results from being released for downstream use. Based on this premise, we evaluate the effectiveness of each uncertainty quantification method as an accuracy indicator using the following four metrics:

\begin{itemize}[leftmargin=1em]
    \item \textbf{AUROC}~\cite{xia2025survey} evaluates the discriminative ability of uncertainty scores to separate correct and incorrect predictions. Specifically, by treating incorrect responses as positive cases and uncertainty scores as ranking signals, AUROC measures the probability that a randomly chosen incorrect response is assigned a higher uncertainty than a randomly chosen correct response. Higher AUROC indicates better discrimination between reliable and unreliable predictions.
    \item \textbf{C-Index}~\cite{steck2007ranking} is a rank-based metric commonly used in survival analysis to assess the consistency between predicted risk scores and observed outcomes. In our setting, where assessment scores are ordinal rather than binary, we use the C-index to evaluate whether responses with larger true errors tend to receive higher uncertainty scores. This extends AUROC to continuous or ordinal error magnitudes by measuring the fraction of concordant pairs between uncertainty and error rankings. Higher C-index indicates better alignment between uncertainty and error severity.
    \item \textbf{AUARC}~\cite{xia2025survey} measures the trade-off between prediction accuracy and rejection rate when selectively abstaining from low-confidence responses. The accuracy--rejection curve plots the accuracy of retained predictions as a function of the fraction of responses rejected based on uncertainty. AUARC summarizes this curve as a single scalar, where higher values indicate that rejecting a small fraction of high-uncertainty responses yields substantial gains in accuracy. This metric reflects the practical utility of uncertainty for selective prediction and human-in-the-loop workflows.
    \item \textbf{AUERC}~\cite{simeone2011shaping} is the error-based counterpart of AUARC, where the error--rejection curve plots the mean absolute error of retained predictions against the rejection rate. This metric evaluates whether high-uncertainty responses indeed correspond to larger errors. Unlike AUARC, lower AUERC values are better, as they indicate that rejecting uncertain responses effectively reduces the remaining prediction error. AUERC is particularly suitable when assessment outputs are ordinal or continuous scores rather than binary correctness labels.
\end{itemize}

\subsubsection{Stability}

All uncertainty quantification methods in this work are derived from stochastic responses generated by LLMs. As a result, the estimated uncertainty values are inherently noisy due to the randomness of the generation process. Although increasing the number of sampled responses can reduce this variance, repeatedly generating a large number of responses for the same grading request is computationally expensive and impractical in real-world deployment. To study this trade-off, we analyze how uncertainty estimates evolve as the response set size n increases, and measure the rate at which the uncertainty values stabilize. Beyond the absolute values, uncertainty is primarily used for ranking responses (e.g., flagging the top fraction of high-uncertainty cases for human review). Therefore, we additionally evaluate the stability of uncertainty-induced rankings using Spearman’s rank correlation coefficient (Spearmanr)~\cite{bishara2012testing} between estimates computed from different response set sizes. Concretely, given a sequence of $N$ sampled responses, we compute uncertainty values incrementally using the first $k$ responses for $k = 2, 3, \ldots, N$, and measure (i) the relative change in uncertainty values between successive steps (from $k$ to $k{+}1$), and (ii) the Spearman rank correlation between the corresponding uncertainty rankings. The final stability score is obtained by averaging these stepwise change ratios and rank correlations across all valid steps. Lower change ratios and higher Spearman correlations indicate more stable uncertainty estimates.

\subsubsection{Correlation}

Although different uncertainty quantification methods assess response uncertainty from diverse perspectives, correlations among these methods often arise due to shared underlying evidence sources (e.g., categorical frequency vs. relational similarity). To provide an overview of the relationships among methods and to help future work avoid selecting redundant measures, we analyze pairwise correlations between methods using the Pearson correlation coefficient.

\begin{table}[!btph]
\vspace{-0.25cm}
\centering
\caption{List of questions evaluated by our benchmark.}
\vspace{-0.25cm}
\label{tab:question_details}
\resizebox{.475\textwidth}{!}{
\begin{tabular}{@{}cccc|cccc@{}}
\toprule
\textbf{Question} & \textbf{Size} & \textbf{Score} & \textbf{Subject} & \textbf{Question} & \textbf{Size} & \textbf{Score} & \textbf{Subject} \\ \midrule
ASAP & 500 & 0-3 & Essay Writing & Semveal & 500 & 0-3 & Mixture \\ \midrule
T1-DCI & 124 & 0-3 & Chemistry & T1-SEP & 124 & 0-3 & Chemistry \\
T2-DCI & 141 & 0-3 & Chemistry & T2-SEP & 141 & 0-3 & Chemistry \\ \midrule
T1-CK & 208 & 0-3 & Math Pedagogy & T1-PCK & 188 & 0-3 & Math Pedagogy \\
T2-CK & 202 & 0-3 & Math Pedagogy & T2-PCK & 188 & 0-3 & Math Pedagogy \\ \midrule
U41Q1 & 159 & 0-4 & Earth Science & U41Q2 & 184 & 0-4 & Earth Science \\
U42Q1 & 198 & 0-4 & Earth Science & U42Q2 & 209 & 0-4 & Earth Science \\ \bottomrule
\end{tabular}}
\end{table}

\subsection{Evaluation Setting}

\subsubsection{Dataset} To provide a comprehensive evaluation, we collect 14 grading questions from five different sources for our experiments, including two open-source datasets and three private datasets. Specifically, the two open-source datasets are the Automated Student Assessment Prize (ASAP) dataset~\cite{hamner2012asap}, which focuses on short-essay grading (150–550 words) written by students in Grades 7–10, and SemEval-2013 Task 7 (SemEval-13-T7)~\cite{dzikovska2013semeval}, which scores student responses to questions collected from in-class exercises, tests, and tutorial dialogues. For the three private datasets, we manually collect two sets of questions from elementary and middle school classes covering chemistry and general science following the 3DLP standard~\cite{he2024school}, and one set from adult learners focusing on pedagogical skills~\cite{copur2022mathematics}. For the open-source datasets, we use the original ground-truth labels to evaluate the correctness of LLM-based automatic grading. For the private datasets, the ground-truth labels are obtained via agreement between two expert human graders. In cases of disagreement, a third grader adjudicates to determine the final label. Based on these labels, we compute the correctness of the automatic assessment results. In Table~\ref{tab:question_details}, we summarize the statistics and characteristics of each dataset and the corresponding grading questions. For each question, we randomly sample 60–150 answers covering all score categories for evaluation.

\subsubsection{Large Language Models}

To ensure the robustness of our conclusions, we benchmark a comprehensive set of mainstream LLMs, including seven closed-source API-based models from three leading vendors, i.e., OpenAI, Anthropic, and Google, and seven open-source models released by research teams worldwide. For the closed-source models, we include multiple model sizes within the same model families to assess scaling effects. For the open-source models, we cover recent representative variants of standard LLMs, including mixture-of-experts models~\cite{zhou2022mixture} and explicitly reasoning-oriented models~\cite{jaech2024openai}. With this broad model coverage, we aim to ensure that our benchmark results generalize across diverse deployment scenarios of automatic assessment. Table~\ref{tab:selected_llms} summarizes the details of all evaluated models.

\input{modellist}

\subsubsection{Generation Settings}


To ensure broad coverage of uncertainty quantification methods under realistic automatic assessment settings, we implement three widely used generation strategies adopted in recent automatic assessment studies~\cite{chu2024llm}:

\begin{itemize}[leftmargin=1em]
    \item \textbf{zero-shot} is the most straightforward way to leverage LLMs for automatic assessment. Given a question, a student answer, and a grading rubric, the LLM is prompted to directly produce a score without any demonstrations or intermediate reasoning steps.
    \item \textbf{zero-shot + Chain of Thought (COT)} extends \textbf{zero-shot} by eliciting intermediate reasoning steps from the LLM. By decomposing the grading process into a step-by-step reasoning trajectory, the model is better able to align with the grading criteria specified in the rubric and to analyze student responses more comprehensively, often resulting in improved grading performance.
    \item \textbf{few-shot + Chain of Thought (COT)} further augments the prompt with exemplar answers corresponding to different score levels. These demonstrations provide concrete interpretations of the abstract criteria in the rubric, helping the model better internalize grading standards and produce more accurate and consistent assessment outcomes.  Specifically, in our evaluation, we include two demonstration examples for each score level.
\end{itemize}

\noindent Finally, to balance computational cost and practical deployment constraints, we set the number of repeated grading samples per answer to five.

\subsection{Effectiveness Result}

\subsubsection{Overall Result}

Due to variations in the number of score levels and the distribution of responses across questions, the scale of uncertainty values produced by the same method can differ substantially across datasets and tasks. To avoid dominance by large-magnitude uncertainty values for specific questions, we rank each uncertainty method within each configuration (model, dataset, and generation strategy) and then aggregate results by averaging ranks across settings. Table~\ref{tab:overall_effect_rank} summarize the aggregated rankings of all uncertainty quantification methods in the four evaluation metrics. We observe that the \textbf{CE} method consistently attains the top rank across different questions, models, and generation strategies. This finding contrasts with prior uncertainty quantification studies that focus primarily on open-ended Q\&A tasks~\cite{xia2025survey}. In addition, other categorical-based methods, including MAR, FSD, and Numset, consistently outperform relation-based methods. This suggests that the discrete output space characteristic of most automatic assessment tasks makes categorical uncertainty measures more compatible with practical deployment. Within the group of relation-based methods, the average ranking follows the order \textbf{JS} < \textbf{NLI} < \textbf{Embed}. This indicates that, despite its simplicity, Jaccard similarity (JS) effectively captures surface-level agreement among responses. In contrast, although embedding- and NLI-based methods leverage more sophisticated pretrained encoders for sentence-pair comparison, they may require additional adaptation or task-specific calibration to achieve optimal performance in uncertainty quantification. Notably, this observation differs from trends reported in Q\&A-centric benchmarks, where more advanced relation graph constructions (e.g., NLI-based similarity) often demonstrate superior performance.

\begin{table}[]
\caption{Average ranks of methods across questions, datasets, and generation strategies on effectiveness metrics.}
\vspace{-0.25cm}
\label{tab:overall_effect_rank}
\resizebox{.45\textwidth}{!}{
\begin{tabular}{@{}ccccc@{}}
\toprule
\textbf{Method} & \textbf{AUROC} & \textbf{C-index} & \textbf{AUARC} & \textbf{AUERC} \\ \midrule
CE & 4.71 & 5.44 & 4.91 & 5.50 \\
FSD & 5.20 & 5.67 & 5.24 & 5.61 \\
MAR & 4.82 & 5.50 & 5.01 & 5.51 \\
Numset & 5.86 & 6.55 & 5.75 & 6.32 \\ \midrule
JS\_NAD & 6.97 & 6.69 & 7.03 & 6.78 \\
JS\_GE & 8.00 & 7.40 & 8.04 & 7.56 \\
JS\_Eigen & 7.87 & 7.42 & 7.84 & 7.47 \\ \midrule
NLI\_NAD & 7.66 & 7.74 & 7.90 & 8.15 \\
NLI\_GE & 8.18 & 8.10 & 8.17 & 8.22 \\
NLI\_Eigen & 8.23 & 8.10 & 8.21 & 8.34 \\
NLI\_DSE & 9.95 & 9.70 & 9.18 & 9.08 \\ \midrule
Embed\_NAD & 8.32 & 8.15 & 8.51 & 8.23 \\
Embed\_GE & 8.95 & 8.62 & 9.00 & 8.57 \\
Embed\_Eigen & 9.03 & 8.69 & 9.07 & 8.61 \\ \bottomrule
\end{tabular}}
\vspace{-.5cm}
\end{table}

\begin{figure*}[!bpth]
    \centering
    \begin{subfigure}[b]{\textwidth}
        \includegraphics[width=\linewidth]{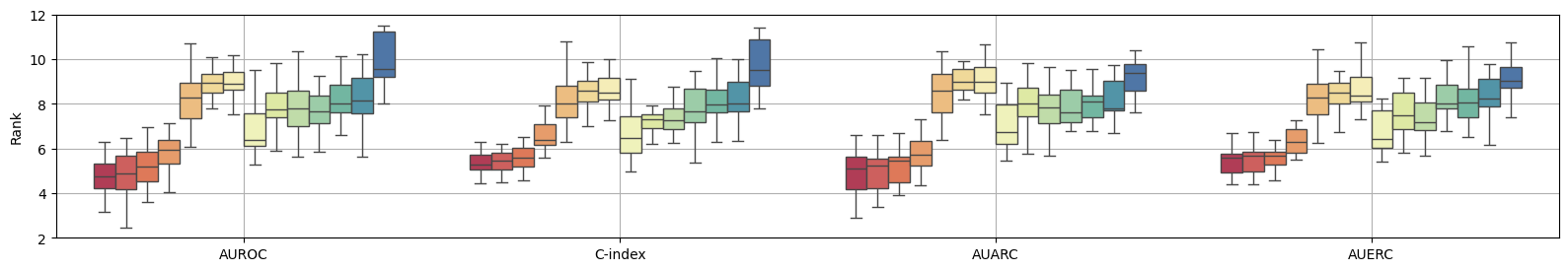}
        \vspace{-0.5cm}
        \caption{Models.}
        \label{fig:model}
    \end{subfigure}
    \begin{subfigure}[b]{\textwidth}
        \includegraphics[width=\linewidth]{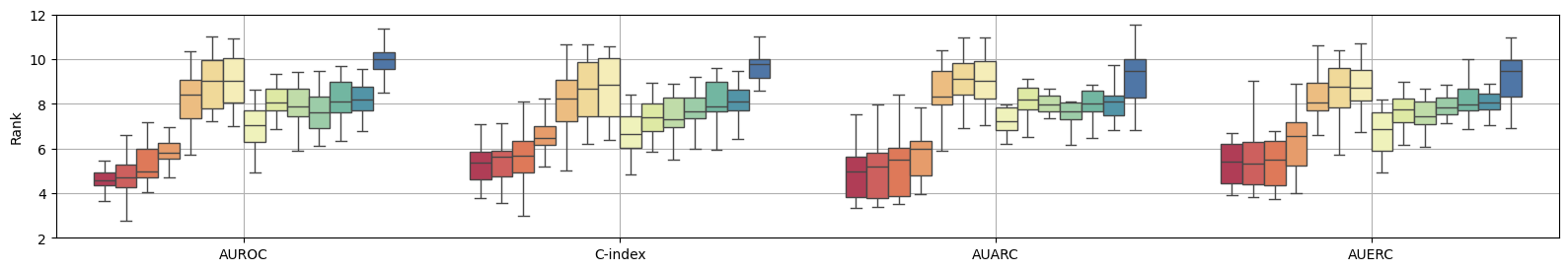}
        \vspace{-0.5cm}
        \caption{Questions.}
        \label{fig:question}
    \end{subfigure}
    \begin{subfigure}[b]{\textwidth}
        \includegraphics[width=\linewidth]{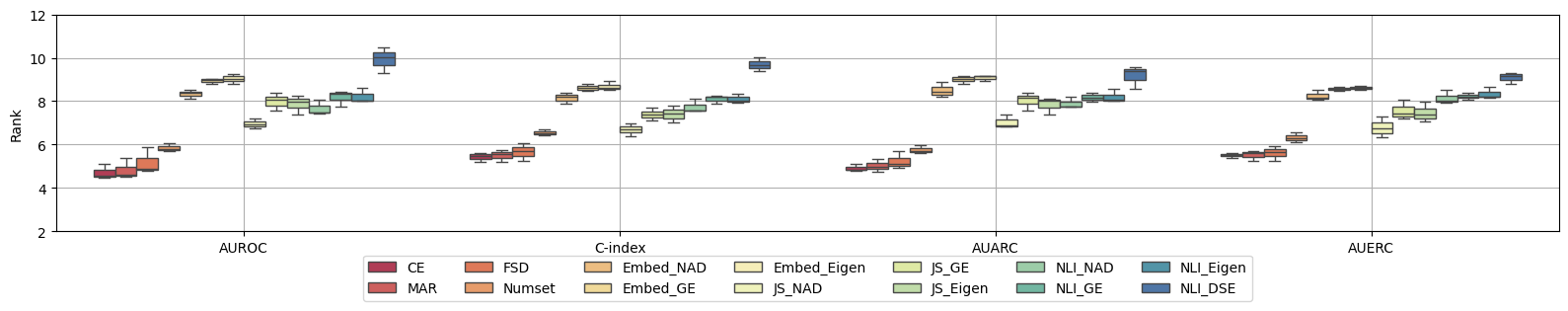}
        \vspace{-0.5cm}
        \caption{Generation strategies.}
        \label{fig:strategy}
    \end{subfigure}
    \vspace{-.5cm}
    \caption{Distribution of various in-group average rank for each model, question and generation strategies variants.}
    \label{fig:perspective_boxplot}
\end{figure*}

\subsubsection{Perspective-specific Discussion}
\label{sec:discuss}

In this section, we examine the robustness of our previously drawn conclusions under variations in questions, models, and generation strategies. Specifically, we compute average ranks for each uncertainty quantification method under each evaluation metric and summarize their distributions across different models, questions, and generation strategies. Figure~\ref{fig:perspective_boxplot} compares these distributions across the three perspectives. From the figure, we make the following observations. First, the variability of method rankings differs across perspectives: generation strategies exhibit the smallest variance, followed by models, while questions show the largest variance. This indicates that method rankings are generally stable across prompting strategies, exhibit moderate variation across models, and are most sensitive to the specific questions being graded. Consequently, while a method may perform consistently under different controls, the optimal choice of uncertainty metric may depend on the particular task or question context. Second, when comparing across methods, categorical-based approaches (e.g., CE, MAR, FSD, and Numset) consistently achieve higher average ranks, with limited overlap with relation-based methods across perspectives. This suggests that, although categorical methods may not always be optimal for specific models or questions, they serve as a strong and reliable default choice in most scenarios. Finally, within the group of relation-based methods, the embedding-based approach exhibits relatively larger variance across perspectives than other relation-based methods on metrics such as AUROC and C-index. This implies that pretrained sentence embeddings are more sensitive to external factors (e.g., dataset characteristics and prompting strategies) and may require more careful calibration or task-specific adaptation when applied to uncertainty quantification.

\begin{figure*}[!btph]
    \centering
    \begin{subfigure}[b]{.32\textwidth}
        \includegraphics[width=\linewidth]{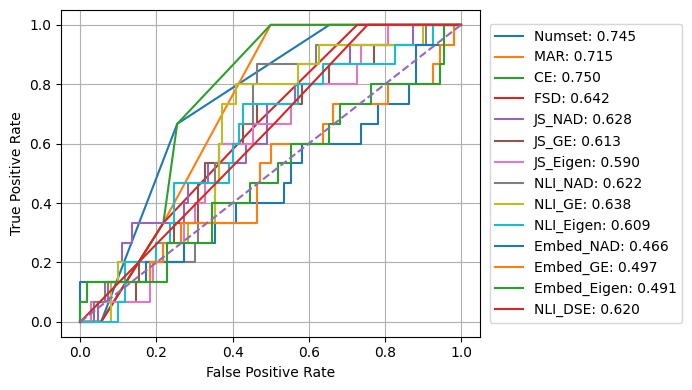}
        \caption{stablelm-2-1-6b-chat (AUROC)}
        \label{fig:asap_stable_roc}
    \end{subfigure}
    \begin{subfigure}[b]{.32\textwidth}
        \includegraphics[width=\linewidth]{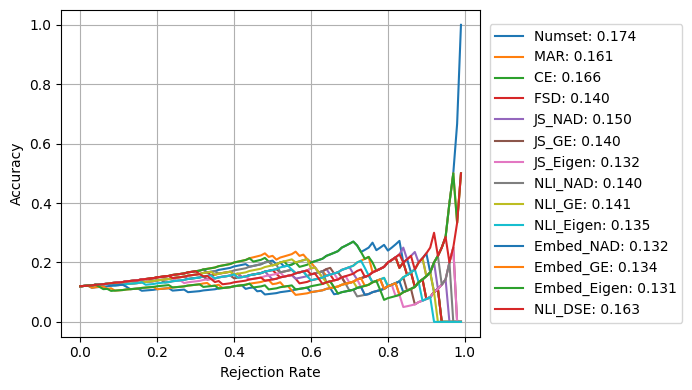}
        \caption{stablelm-2-1-6b-chat (AUARC)}
        \label{fig:asap_stable_arc}
    \end{subfigure}
    \begin{subfigure}[b]{.32\textwidth}
        \includegraphics[width=\linewidth]{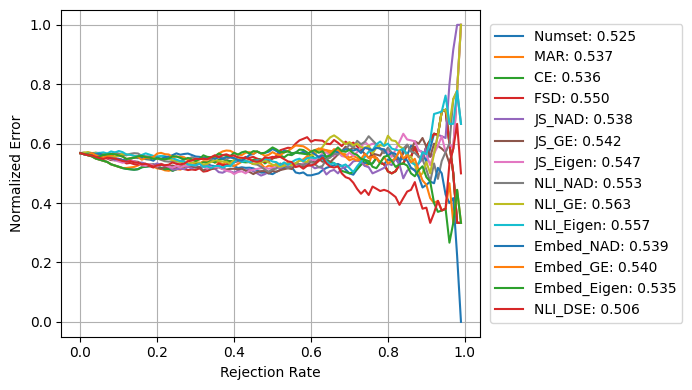}
        \caption{stablelm-2-1-6b-chat (AUERC)}
        \label{fig:asap_stable_erc}
    \end{subfigure}
    \begin{subfigure}[b]{.32\textwidth}
        \includegraphics[width=\linewidth]{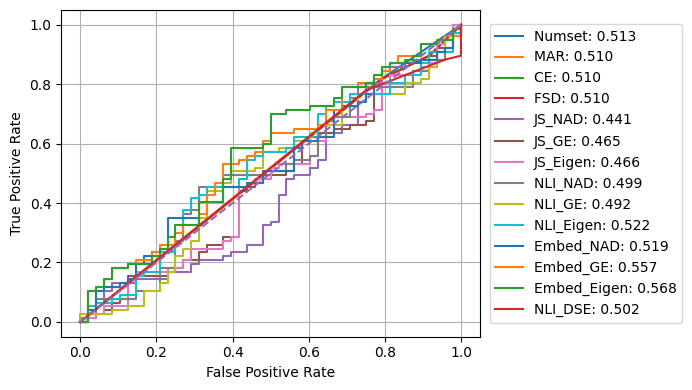}
        \caption{gemini-3-flash (AUROC)}
        \label{fig:asap_gemini_roc}
    \end{subfigure}
    \begin{subfigure}[b]{.32\textwidth}
        \includegraphics[width=\linewidth]{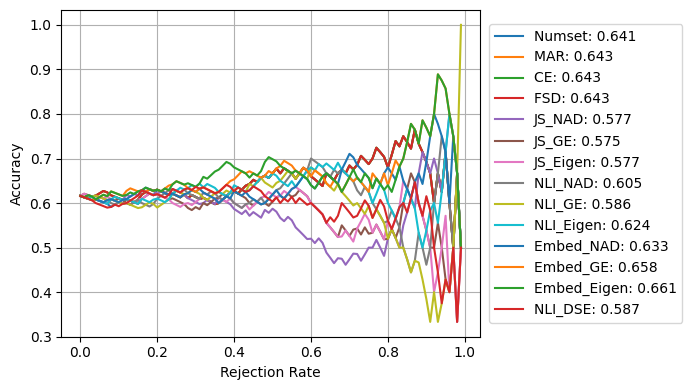}
        \caption{gemini-3-flash (AUARC)}
        \label{fig:asap_gemini_arc}
    \end{subfigure}
    \begin{subfigure}[b]{.32\textwidth}
        \includegraphics[width=\linewidth]{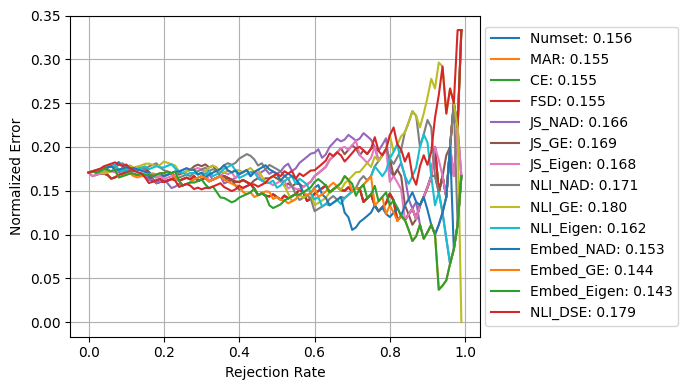}
        \caption{gemini-3-flash (AUERC)}
        \label{fig:asap_gemini_erc}
    \end{subfigure}
    \vspace{-0.25cm}
    \caption{Method Comparison between stablelm and gemini over different metrics on the ASAP question.}
    \label{fig:case_asap}
\end{figure*}

\begin{figure*}[!btph]
    \centering
    \begin{subfigure}[b]{.32\textwidth}
        \includegraphics[width=\linewidth]{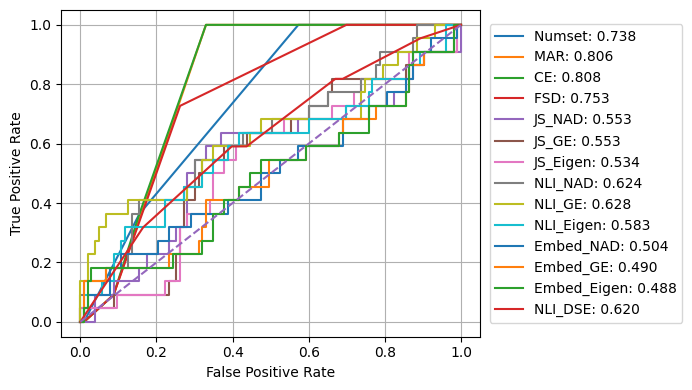}
        \caption{stablelm-2-1-6b-chat (AUROC)}
        \label{fig:semeval_stable_roc}
    \end{subfigure}
    \begin{subfigure}[b]{.32\textwidth}
        \includegraphics[width=\linewidth]{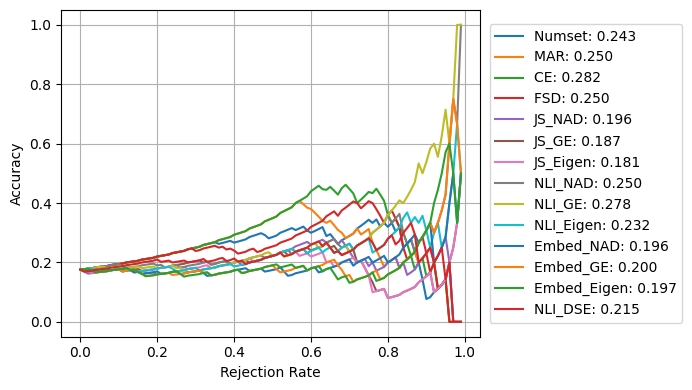}
        \caption{stablelm-2-1-6b-chat (AUARC)}
        \label{fig:semeval_stable_arc}
    \end{subfigure}
    \begin{subfigure}[b]{.32\textwidth}
        \includegraphics[width=\linewidth]{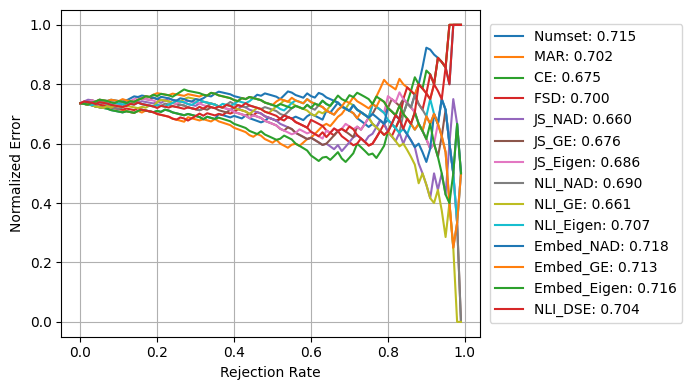}
        \caption{stablelm-2-1-6b-chat (AUERC)}
        \label{fig:semeval_stable_erc}
    \end{subfigure}
    \begin{subfigure}[b]{.32\textwidth}
        \includegraphics[width=\linewidth]{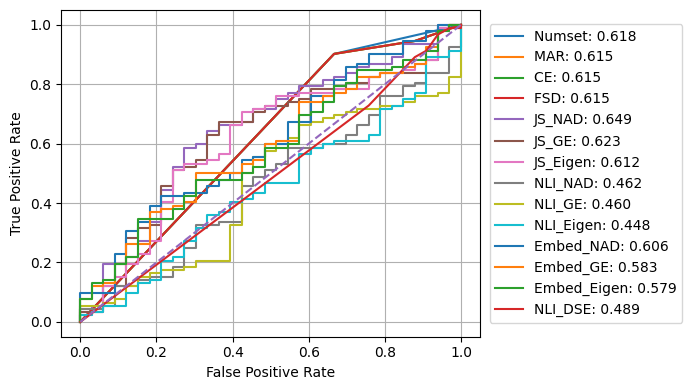}
        \caption{gemini-3-flash (AUROC)}
        \label{fig:semeval_gemini_roc}
    \end{subfigure}
    \begin{subfigure}[b]{.32\textwidth}
        \includegraphics[width=\linewidth]{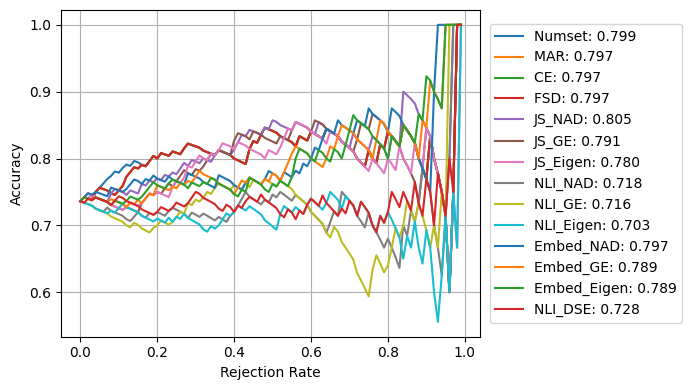}
        \caption{gemini-3-flash (AUARC)}
        \label{fig:semeval_gemini_arc}
    \end{subfigure}
    \begin{subfigure}[b]{.32\textwidth}
        \includegraphics[width=\linewidth]{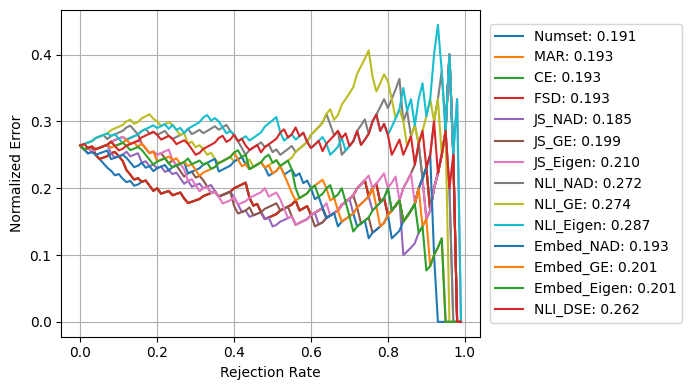}
        \caption{gemini-3-flash (AUERC)}
        \label{fig:semeval_gemini_erc}
    \end{subfigure}
    \caption{Method Comparison between stablelm and gemini over different metrics on the SemEval question.}
    \vspace{-0.25cm}
    \label{fig:case_semeval}
\end{figure*}

\subsubsection{Case Studies}

To provide the specific evidence to support our conclusions in section~\ref{sec:discuss} we conduct case studies on the two open-source datasets ASAP and SemEval. We further select two representative models, stablelm-2-1-6b-chat and gemini-3-flash-preview, which exhibit the lowest and highest average grading performance across questions, respectively, and evaluate them under the best-performing generation strategy (few-shot + CoT). Specifically, we illustrate the ROC, accuracy–rejection (ARC), and error–rejection (ERC) curves for both models on ASAP (Figure~\ref{fig:case_asap}) and SemEval (Figure~\ref{fig:case_semeval}). By comparing AUROC, AUARC, and AUERC values across models and datasets, we observe that the effectiveness of uncertainty methods depends on the underlying model. For instance, categorical-based methods consistently outperform relation-based methods for stablelm-2-1-6b-chat, whereas this trend is reversed for gemini-3-flash-preview. This observation aligns with our findings in the previous section that method rankings are influenced by factors such as model quality and dataset characteristics. Further inspection of the generated grading responses reveals that higher-quality outputs from gemini-3-flash-preview lead to more semantically coherent and informative rationales, making relation-based uncertainty measures more reliable. In contrast, noisier and less accurate responses from lower-performing models limit the effectiveness of relational analysis, favoring categorical-based uncertainty measures. Accordingly, we recommend prioritizing categorical-based methods when model performance is low, and shifting to relation-based methods once grading accuracy surpasses a reasonable threshold. Finally, comparing method rankings for the same model across the two datasets shows that lower-performing models exhibit relatively stable gains from categorical-based methods, whereas for higher-performing models, the effectiveness of relation-based methods is more sensitive to how the relation graph is constructed. This suggests that, when working with high-performing LLMs, careful design of the relation graph construction strategy is crucial for the effectiveness of relational uncertainty quantification.

\subsection{Stability Result}

\subsubsection{General Result}

Table~\ref{tab:overall_stable_rank} reports the stability evaluation results across questions, models, and generation strategies. Consistent with the effectiveness analysis, we compare methods based on their aggregated ranks over the change-ratio and Spearman correlation metrics. In contrast to effectiveness results, categorical-based methods do not consistently dominate relation-based methods in terms of stability. Notably, relation-based approaches grounded in graph properties—such as NAD- and GE-based variants (e.g., JS\_NAD and JS\_GE)—exhibit more stable uncertainty estimates as the number of sampled responses increases. It suggests that while categorical methods may achieve stronger overall effectiveness in ranking performance, certain relation-based methods provide superior robustness with respect to sampling variability. Meanwhile, we observe that the categorical method CE, although not the most stable, performs competitively and remains close to the top-tier relation-based methods in terms of stability.

\begin{table}[]
\caption{Average ranks of methods across questions, datasets, and generation strategies on stability metrics.}
\vspace{-0.25cm}
\label{tab:overall_stable_rank}
\resizebox{.475\textwidth}{!}{
\begin{tabular}{@{}ccc|ccc@{}}
\toprule
\textbf{Method} & \textbf{Delta} & \textbf{Spearmanr} & \textbf{Method} & \textbf{Delta} & \textbf{Spearmanr} \\ \midrule
CE & 5.30 & 4.51 & NLI\_NAD & 8.50 & 7.01 \\
FSD & 7.69 & 6.58 & NLI\_GE & 8.05 & 9.93 \\
MAR & 8.65 & 5.37 & NLI\_Eigen & 12.10 & 10.56 \\
Numset & 2.47 & 6.26 & NLI\_DSE & 13.57 & 12.84 \\ \midrule
JS\_NAD & 2.84 & 3.86 & Embed\_NAD & 6.00 & 4.92 \\
JS\_GE & 1.89 & 7.23 & Embed\_GE & 5.02 & 7.89 \\
JS\_Eigen & 10.91 & 7.37 & Embed\_Eigen & 11.95 & 8.20 \\ \bottomrule
\end{tabular}}
\vspace{-.5cm}
\end{table}

\subsubsection{Perspective-specific Discussion}

\begin{figure}[!btph]
  \centering
  \includegraphics[width=\linewidth]{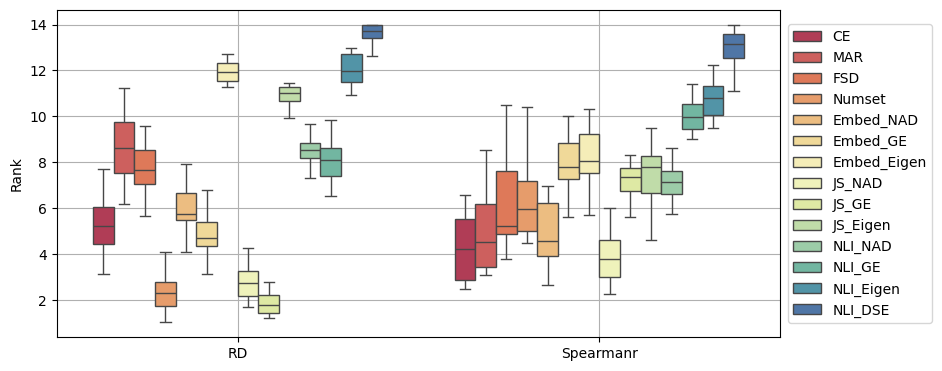}
  \includegraphics[width=\linewidth]{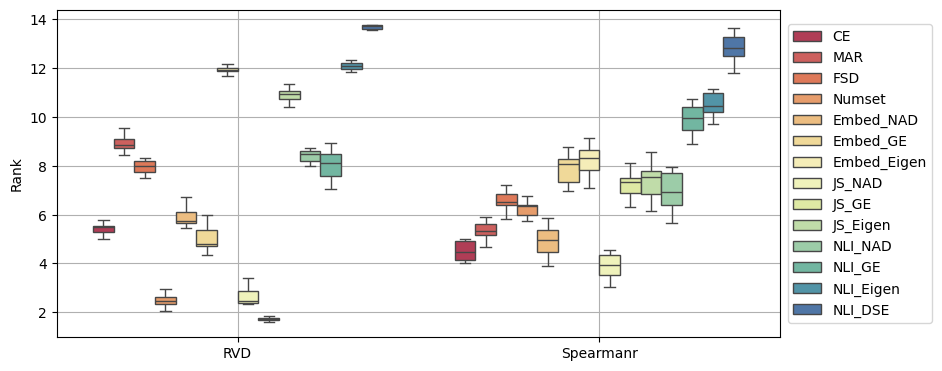}
  \includegraphics[width=\linewidth]{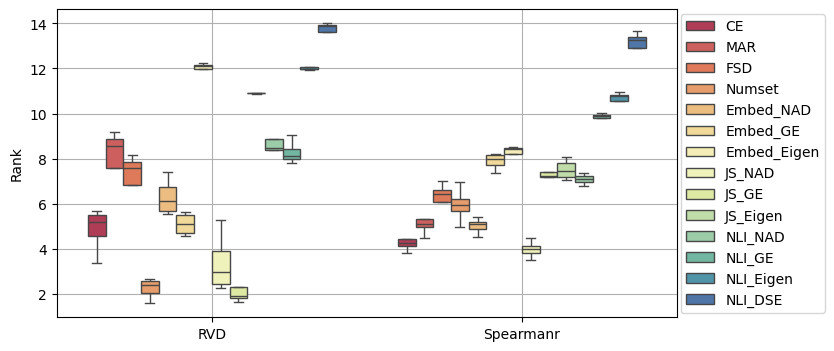}
  \caption{Distribution of various in-group average rank in stability across model, question and strategy.}
  \label{fig:stable_box}
\end{figure}

Similar to Section~\ref{sec:discuss}, we analyze the robustness of our conclusions regarding stability rankings under variations in three factors: model, question, and generation strategy. Figure~\ref{fig:stable_box} presents the distribution of stability ranks for each method across these factors. From the figure, we observe that, in contrast to the four effectiveness metrics, stability metrics are relatively insensitive to variations in questions and generation strategies. However, model choice significantly influences the relative stability rankings of different methods. This observation indicates that stability should be re-evaluated when applying uncertainty quantification methods to new models, as method robustness does not necessarily transfer across model families.

\subsection{Correlation Result}

\subsubsection{General Result}
\begin{figure}
    \centering
    \begin{subfigure}[b]{.23\textwidth}
        \includegraphics[width=\linewidth]{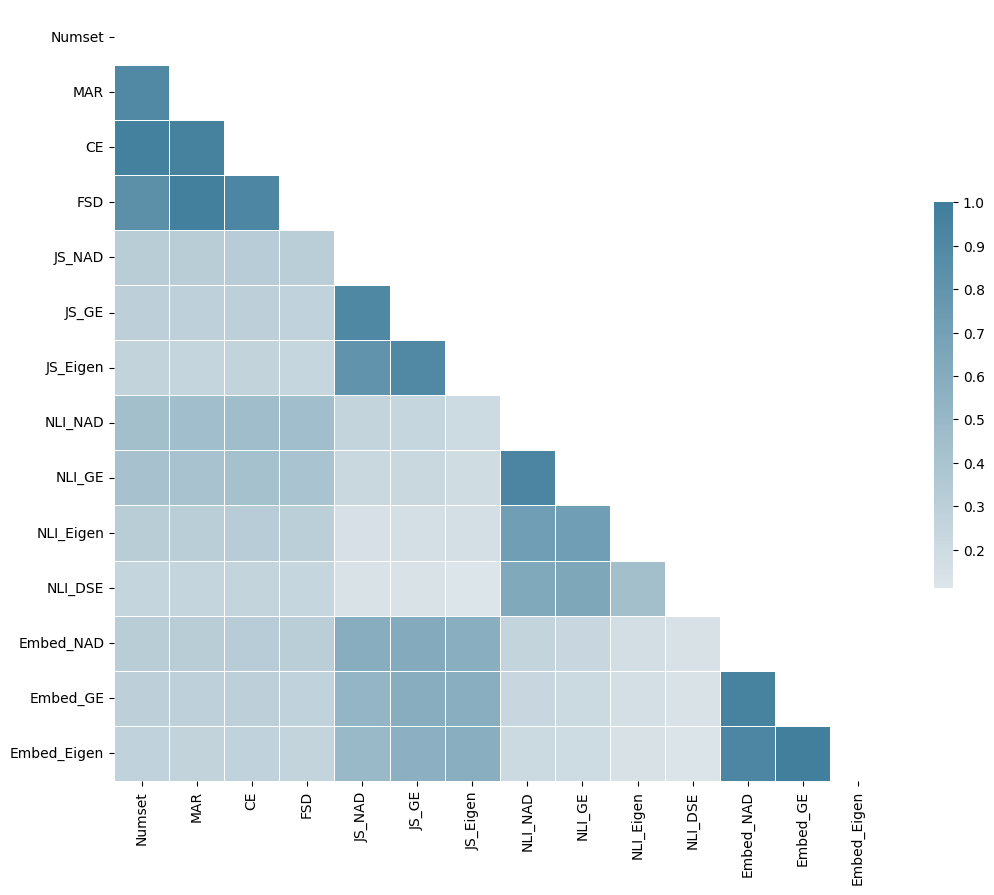}
        \caption{General mean values.}
        \label{fig:overall_mean}
    \end{subfigure}
    \begin{subfigure}[b]{.23\textwidth}
        \includegraphics[width=\linewidth]{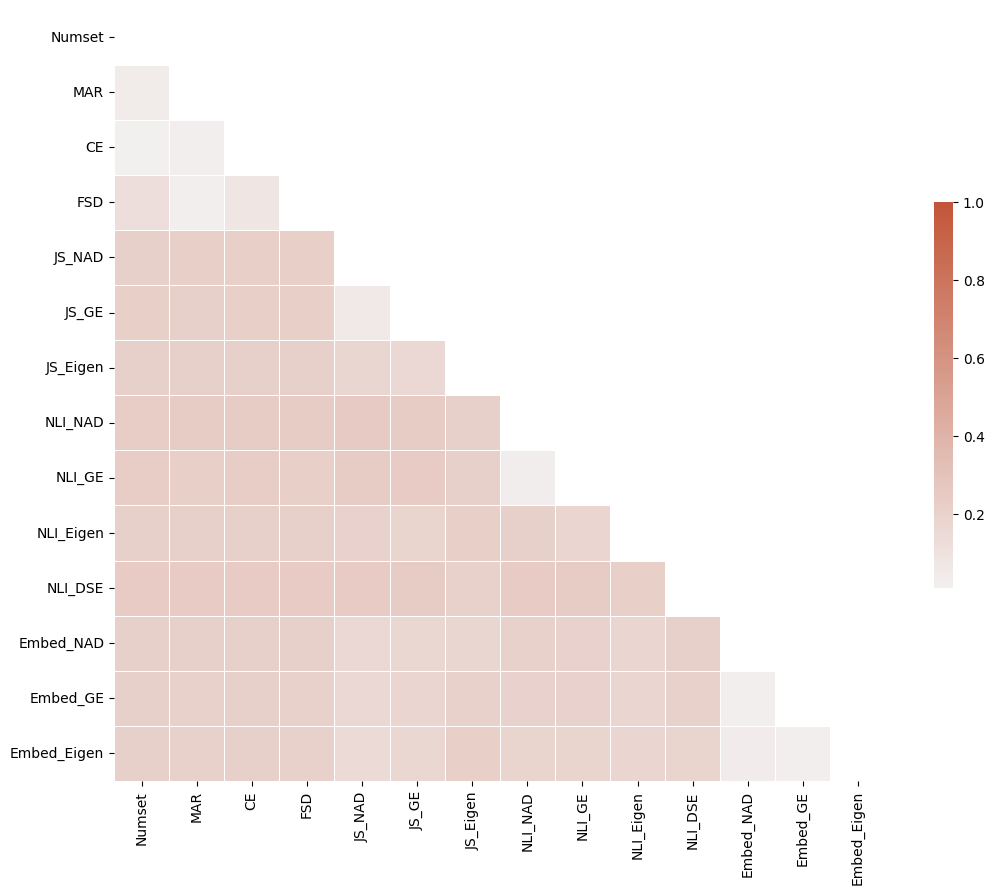}
        \caption{Variance over models.}
        \label{fig:model_var}
    \end{subfigure}
    \begin{subfigure}[b]{.23\textwidth}
        \includegraphics[width=\linewidth]{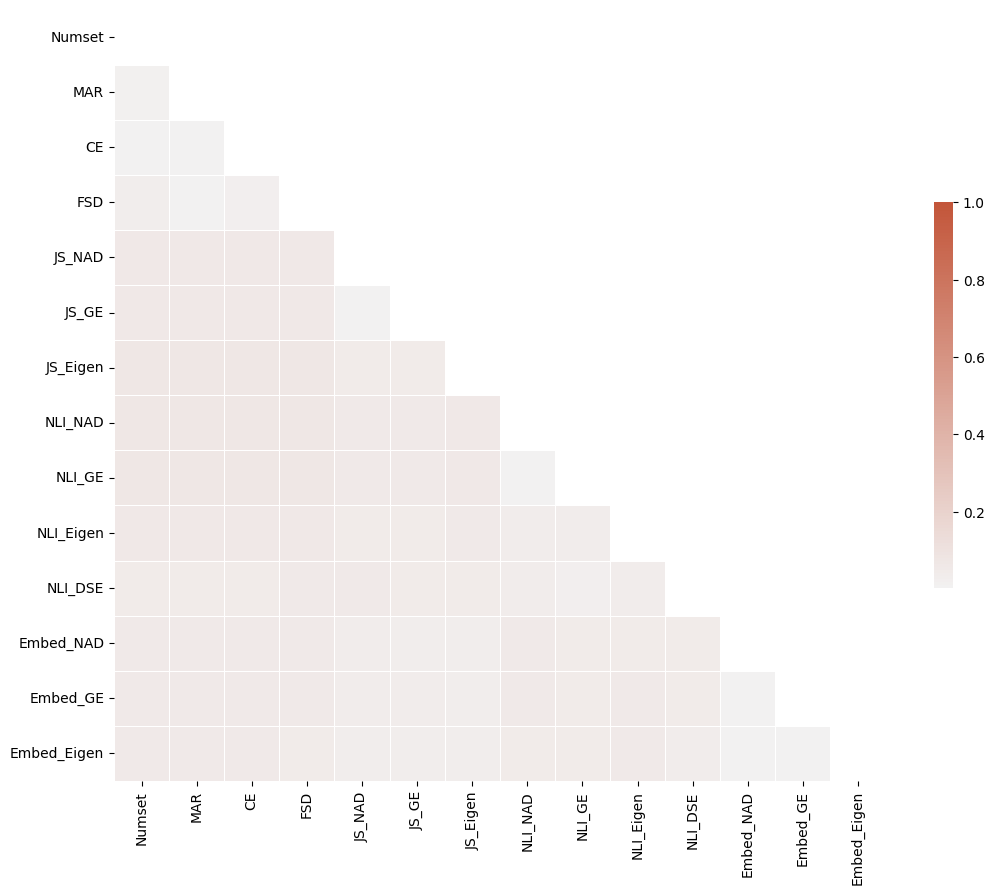}
        \caption{Variance over questions.}
        \label{fig:data_var}
    \end{subfigure}
    \begin{subfigure}[b]{.23\textwidth}
        \includegraphics[width=\linewidth]{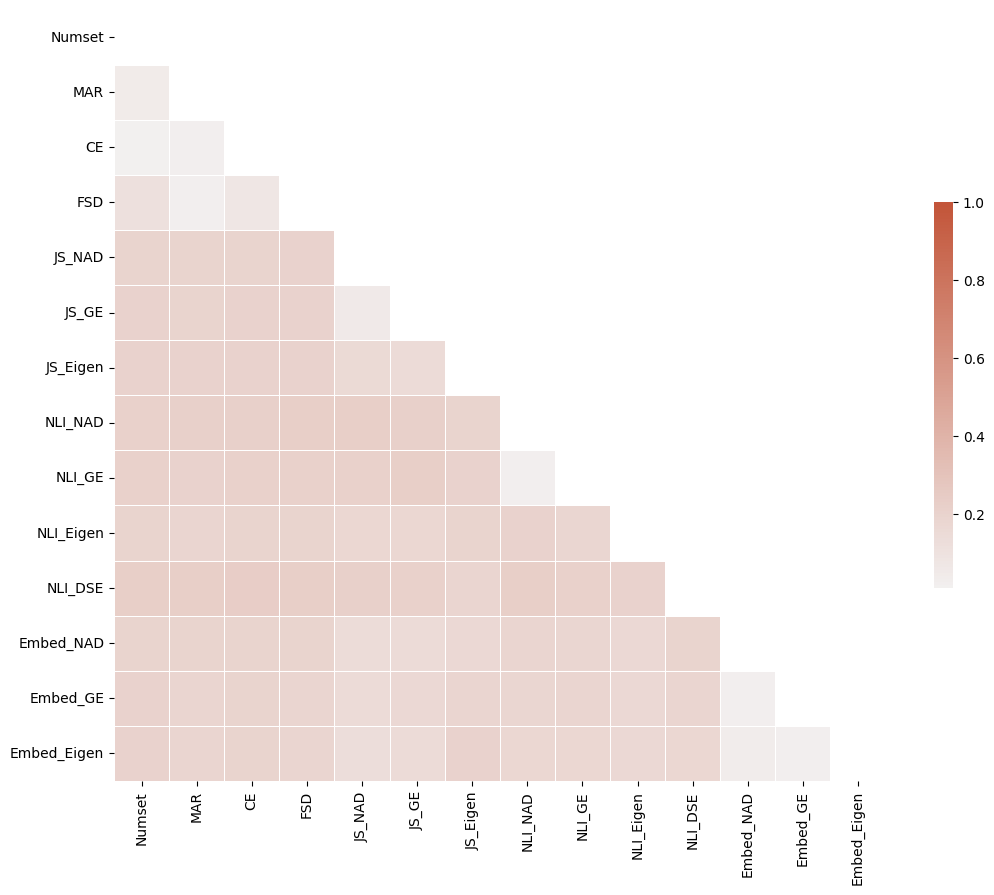}
        \caption{Variance over strategics.}
        \label{fig:strategy_var}
    \end{subfigure}
    \caption{Mean and variance of aggregated pairwise correlation values among methods across questions, models, and generation strategies.}
    \label{fig:overall_corr}
    \vspace{-0.5cm}
\end{figure}

To analyze correlations among methods, we compute the Pearson correlation coefficient for each pair of uncertainty quantification methods and aggregate these correlations across questions, models, and generation strategies by averaging. The resulting aggregated correlation matrix is shown in Figure~\ref{fig:overall_mean}. From the figure, we observe strong correlations among categorical-based methods, reflecting their reliance on similar frequency-based evidence. Among relation-based methods, those that share the same graph construction strategy exhibit particularly high correlations. In addition, methods derived from embedding-based and Jaccard similarity graphs show moderate correlation, as both capture semantic relatedness between responses, albeit at different representational levels.

\subsubsection{Perspective-specific Discussion}
Similar to our analyses of effectiveness and stability, we examine the robustness of our correlation findings under variations in questions, models, and generation strategies. As shown in Figures~\ref{fig:data_var}, Figure~\ref{fig:model_var} and Figure~\ref{fig:strategy_var}, although variability exists across different models, questions, and strategies, the magnitude of this variance is relatively small (typically < 0.2) compared to the average correlation values (generally > 0.5). This indicates that the overall correlation patterns reported in our general results are robust to these sources of variation. 

%% file: modellist.tex
\begin{table}[]
\caption{List of LLMs evaluated by our benchmark.}
\label{tab:selected_llms}
\resizebox{.475\textwidth}{!}{
\begin{tabular}{@{}cc@{}cc@{}}
\toprule
\multicolumn{3}{c}{\textbf{Proprietary Models}} \\ 
\midrule
\textbf{Model} & \textbf{Model ID} & \textbf{Scale} \\ 
\midrule
Calude Haiku 4.5\cite{anthropic2025haiku45} & claude-haiku-4-5-20251001 & / \\
Calude Sonnet 4.5\cite{anthropic2025claude45} & claude-sonnet-4-5-20250929 & / \\
Gemini 2.5 Pro~\cite{team2023gemini} & gemini-2.5-pro & / \\
Gemini 3 Flash~\cite{team2023gemini} & gemini-3-flash-preview & / \\
GPT-4o mini~\cite{Hurst2024GPT4oSC} & gpt-4o-mini & / \\
GPT-5 ~\cite{singh2025openai} & gpt-5 & / \\
GPT-5 nano~\cite{singh2025openai} & gpt-5-nano & / \\
\midrule

\midrule
\multicolumn{3}{c}{\textbf{Open-source Models}} \\ 
\midrule
\textbf{Model} & \textbf{Model ID} & \textbf{Scale} \\ 
\midrule
Qwen3 4B Instruct~\cite{qwen3technicalreport} & Qwen/Qwen3-4B-Instruct-2507 & 4B \\
Stable LM 2 1.6B Chat~\cite{StableLM-2-1.6B} & 
stabilityai/stablelm-2-1-6b-chat & 6B \\
Llama 3.1 8B Instruct~\cite{grattafiori2024llama} & meta-llama/Llama-3.1-8B-Instruct & 8B \\
Falcon3 10B Instruct~\cite{Falcon3} & tiiuae/Falcon3-10B-Instruct & 10B \\
Ministral 3 14B Instruct~\cite{liu2026ministral} & mistralai/Ministral-3-14B-Instruct-2512 & 14B \\
DeepSeek R1 Distill - Qwen 32B~\cite{deepseekai2025deepseekr1incentivizingreasoningcapability} & deepseek-ai/DeepSeek-R1-Distill-Qwen-32B & 33B \\
Mixtral 8x7B Instruct & mistralai/Mixtral-8x7B-Instruct-v0.1 & 47B \\
\bottomrule
\end{tabular}}
\end{table}

%% file: conclusion.tex
We present a benchmark of uncertainty quantification methods for LLM-based automatic assessment, and emphasize the reliability of prediction by investigating the stability of the uncertainty metrics. Although uncertainty estimation has been studied in many fields, the automated assessment of LLMs is still underexplored due to the ordinal output and rubric-driven setting. To address this issue, we systematically evaluate a diverse set of UQ metrics across datasets of different subjects, LLMs of different scales,   and prompt strategies to elicit assessment responses. We analyze the effectiveness, stability, and the relations between the UQ methods. Our analysis results reveal several insightful patterns. The categorical-based methods demonstrate strong effectiveness in probing uncertainty in assessment judgment; as for relation-based methods, although the performance is not so competitive, their robustness to sampling variation is superior to support a stable prediction. This phenomenon shows a trade-off between assessment discriminative ability and stability. The inter-metric analyses show clear redundancy within method families and provide practical guidance for selecting complementary uncertainty metrics to achieve better uncertainty modeling efficiently. Overall, our findings highlight that no UQ metric is universally optimal, and selecting the proper UQ metric should take the LLM's capability, assessment tasks, and the settings into consideration. We hope this benchmark facilitates principled evaluation of uncertainty estimation for auto-assessment tasks and supports the development of reliable and practical LLM-based assessment systems. 